\documentclass{article}

     \PassOptionsToPackage{numbers, compress}{natbib}

    \usepackage[final]{neurips_2020}

\usepackage[utf8]{inputenc} 
\usepackage[T1]{fontenc}    
\usepackage{hyperref}       
\usepackage{url}            
\usepackage{booktabs}       
\usepackage{amsfonts}       
\usepackage{amsmath}
\usepackage{epsfig}         
\usepackage{comment}
\usepackage{todonotes}
\usepackage{nicefrac}       
\usepackage{microtype}      
\usepackage[ruled,linesnumbered,noend]{algorithm2e}
\usepackage{enumitem}
\usepackage{graphicx}
\usepackage{tabulary}
\usepackage{multirow}
\usepackage{wrapfig}

\title{Targeted Query-based Action-Space Adversarial Policies on Deep Reinforcement Learning Agents}

\author{%
  Xian Yeow Lee, Yasaman Esfandiari, Kai Liang Tan, Soumik Sarkar \\
  Department of Mechanical Engineering\\
  Iowa State University\\
  Ames, IA 50011 \\
  \texttt{\{xylee, yasesf, kailiang, soumiks\}@iastate.edu} 
}

\begin{document}

\maketitle

\begin{abstract}
Advances in computing resources have resulted in the increasing complexity of cyber-physical systems (CPS). As the complexity of CPS evolved, the focus has shifted from traditional control methods to deep reinforcement learning-based (DRL) methods for control of these systems. This is due to the difficulty of obtaining accurate models of complex CPS for traditional control. However, to securely deploy DRL in production, it is essential to examine the weaknesses of DRL-based controllers (policies) towards malicious attacks from all angles. In this work, we investigate targeted attacks in the action-space domain, also commonly known as actuation attacks in CPS literature, which perturbs the outputs of a controller. We show that a query-based black-box attack model that generate optimal perturbations with respect to an adversarial goal can be formulated as another reinforcement learning problem. Thus, such an adversarial policy can be trained using conventional DRL methods. Experimental results showed that adversarial policies which only observe the nominal policy's output generate stronger attacks than adversarial policies that observe the nominal policy's input and output. Further analysis reveals that nominal policies whose outputs are frequently at the boundaries of the action space are naturally more robust towards adversarial policies. Lastly, we propose the use of adversarial training with transfer learning to induce robust behaviors into the nominal policy, which decreases the rate of successful targeted attacks by half.
\end{abstract}

\section{Introduction}

Learning-based approaches, specifically deep learning, are increasingly being applied to dynamical systems and control. This trend is especially prevalent in cyber-physical systems (CPS), which are capable of generating an enormous amount of data from monitoring systems and sensor readings. Since deep learning models (deep neural networks) are trained in an end-to-end fashion by leveraging the information-rich data, these methods have gained traction for solving various tasks. Examples of deep learning being applied in CPS include fault and anomaly detection~\cite{luo2020deep, mozaffari2020learning}, system monitoring~\cite{lee2020automated}, and supporting human-machine interaction procedures~\cite{lore2016deep,liu2018deep}.

Deep Reinforcement Learning (DRL), which combines deep neural networks as function approximators with reinforcement learning algorithms for sequential decision making have also enjoyed tremendous breakthroughs in recent years, due to algorithmic and computing resource advancements. For example, DRL has successfully been applied to problems such as inverse design~\cite{popova2018deep, lee2019case}, traffic resource management~\cite{jang2019simulation, tan2019deep}, recommendation systems~\cite{10.1145/3178876.3185994} and even image processing~\cite{furuta2019pixelrl}. Within the area of CPS, RL~\footnote{The terms DRL and RL are used interchangeably in this manuscript.} have mainly been applied in place of traditional controllers, where the RL agent makes decisions based on a set of observations (e.g., sensor readings, states of the system), and the decisions get converted into physical actions. While traditional control methods are well-developed, extensive knowledge of the system's behavior is often required to design a suitable controller. As such, traditional control methods are often challenging to scale to complex, high-dimensional systems. On the other hand, RL methods do not require knowledge of the system but only requires access to the actual system or a surrogate of the system to learn a good control policy. Subsequently, a control policy with complex behaviors can emerge from training an RL agent as a CPS controller. More importantly, obtaining a highly accurate model for real-world, complex CPS is often non-trivial. Therefore, RL-approaches which leverage the abundance of data to learn a control policy which can be adapted to real systems is extremely useful. Various case studies have proposed using RL as controllers for complex, high-dimensional CPS, such as robotic manipulation and navigation~\cite{7989385, 7989381}.

Unfortunately, RL-based control has also been proven to be vulnerable to a host of malicious adversarial attacks~\cite{chen2019adversarial, ilahi2020challenges}. These attacks seek to inject perturbations of various forms through the multiple channels in which the controller interacts with the environment. In CPS applications, especially safety-critical systems, the cost of such a successful attack on the RL controller can be extremely high. Thus, it is vital to understand the various mechanisms in which an RL controller can be attacked to mitigate such scenarios. In this paper, we focus on targeted action-space attacks where the perturbations are injected into the RL-controller's output signal. Since action-space attacks (also known as actuation attacks) are relatively less studied in the context of RL-based controls, this work serves to fill in such a gap. Specifically, we contribute to the following in this paper: We propose a query-based \textit{black-box} DRL-based targeted attack model that perturbs another RL agent to an adversarially-defined goal. We demonstrate that full-state information is not required for to learn a successful attack. Instead, only actuation information and information related to the adversarial goal is sufficient for learning effective attacks. Furthermore, we study the characteristics of the nominal policies and empirically show that policies that favor extreme actions are naturally more robust towards attacks. Additionally, we study the efficacy of different adversarial training schemes and conclude that for our proposed attack model, a combination of transfer learning and adversarial training is required to robustify nominal policies against such attacks.

\section{Related Works}
\label{sec:related_works}

The vulnerability of control systems to adversarial attacks has been shown by many research studies with different applications. The security concerns for control systems has been discussed as early as $1989$ when attack detection and system recovery were proposed in~\cite{7501620} and~\cite{basseville1993detection}. As CPS have specific vulnerabilities dependant on their structure, the general approach is to analyse the adverse effect of particular attacks on  specific systems. For example, in distributed multi-agent systems, there exists attacks which disrupt the states of the system without changing the system's outputs (stealthy attacks), for which several algorithms have been proposed~\cite{vamvoudakis2014detection, guo2016optimal, 10.5555/3172077.3172414}.  

Reinforcement learning agents have recently been found susceptible to attacks as well. Their vulnerability has been discussed in~\cite{DBLP:journals/corr/BehzadanM17, s2017adversarial}, where the authors proposed a method to attack a DRL agent's state-space uniformly at each every time step. Several other researchers have also proposed state-space attack and defence strategies for DRL agents~\cite{10.5555/3172077.3172414, DBLP:journals/corr/abs-1712-03632,sun2020stealthy, mandlekar2017adversarially, havens2018online}. While the above studies focus on white-box attacks,~\cite{russo2019optimal} attempted to find optimal black-box attacks and found that smooth policies are naturally more resilient. Also,~\cite{zhao2020blackbox} used sequence-to-sequence models to predict a sequence of actions that a trained agent will take in the future and showed that the black-box attacks can make a trained agent malfunction after a specific time delay. Another study by~\cite{lee2020spatiotemporally} also proved the vulnerability of RL agents to action-space attacks by framing the problem as a constrained-optimization problem of minimizing the total reward by taking into account the dynamics of the agent. Meanwhile,~\cite{tessler2019action} proposed that action-space robustness is a new criteria for real world use of RL and devised an algorithm which makes RL agents more robust to action-space noise. ~\cite{tan2020robustifying} also adversarially trained DRL agents using Projected Gradient Sign~(\textit{PGD}) method~\cite{madry2017towards} to provide action-space robustness. Another branch of studying the robustness of DRL agent focuses on adversarial policies in which another DRL agent is introduced in the system to learn an adversarial policy which makes the nominal DRL agent robust both in natural and adversarial environments~\cite{mandlekar2017adversarially}. Additionally,~\cite{gleave2019adversarial} used adversarial policy training methods in a multi-agent system to train an adversarial agent to attack the nominal agent. In this paper, we focus on a similar adversarial policy training method to train a policy to mount targeted attacks on the nominal agent in the action-space.

\section{Methodology}

\subsection{Deep Reinforcement Learning}

In this section, we provide a brief background on DRL algorithms.
Specifically, we focus on model-free policy-iteration based methods that are tailored for continuous control. Let $s_{t}$ and $a_{t}$ denote the (continuous, possibly high-dimensional) state and action at time $t$ respectively. Given an $a_{t}$ that is selected according to policy $\pi$ at $s_{t}$, a corresponding reward signal, $r_t$ is computed using the reward function, $r_{t} = R(s_{t}, a_{t})$. The reward function is typically predefined and embedded in the environment and remains unknown to the RL agent. The RL agent recursively interacts with the environment with $a_{t}$ and transitions into the next state, $s_{t+1} = E(s_{t}, a_{t})$, until a defined horizon $t = T$, or a completion condition is met. Interacting with the environment creates a sequence of state-action vectors, also known as a trajectory $\tau = (s_{t}, a_{t}, s_{t+1}, a_{t+1}, ...)$. The goal of the RL agent is to learn an optimal policy $\pi^{*}$ so that the discounted cumulative future reward of the entire trajectory, $R(\tau) = \sum_{t=0}^{T} \gamma^{t} r_{t}$, is maximized. The discount factor, $\gamma \in (0,1)$, allows the RL agent to either favor short-term rewards or maximize potentially larger rewards in the future. 

Policy-based methods directly learns a state-action mapping $\pi$ that maximizes the expected return $J(\pi) = \mathbb{E}[R(\tau)]$. Here, $\pi$ is also represented with a deep neural network parameterized by $\theta$, where $\theta$ is optimized using gradient ascent, i.e.:
\begin{equation}
	\theta_{k+1} = \theta_{k} + \alpha \nabla_{\theta} J(\pi_{\theta}) |_{\theta_{k}}
\end{equation}
In practice, the policy returns a distribution over actions via a representation of a parameterized distribution (e.g., Gaussian), where the optimal parameters of the distribution ($\mu$, $\sigma$) are estimated by the neural network. The optimal action $a^{*}_t$ is then sampled from the distribution. 
\begin{equation}
	a_{t}^{*} \sim \pi^{*}_{\theta}, \ \ \pi^{*} = arg\max_{\pi} \mathbb{E}[R(\tau)]
\end{equation}
Variants of policy optimization algorithms used in DRL include Proximal Policy Optimization (PPO)~\cite{schulman2017proximal} and Asynchronous Methods for Deep Reinforcement Learning~\cite{mnih2016asynchronous}. 

\subsection{Targeted Adversarial Perturbations}

\begin{wrapfigure}{r}{0.45\textwidth}
  \begin{center}
    \includegraphics[width=0.45\textwidth, clip, trim={3.2in 1.29in 2.6in 1.13in }]{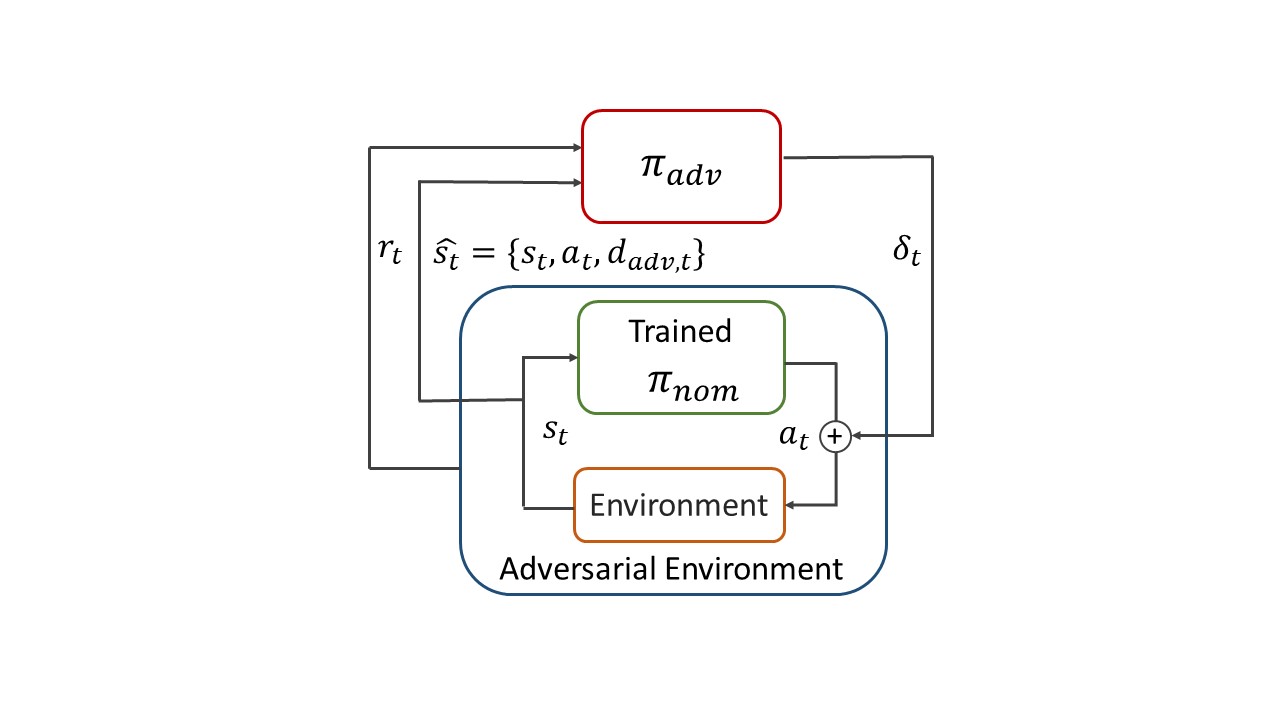}
  \end{center}
  \caption{Overview of our proposed framework for training an adversarial policy, $\pi_{adv}$ to mount targeted attacks in the action-space domain on a trained nominal policy $\pi_{nom}$. }
\end{wrapfigure}
In the following sections, we refer to the RL agent subjected to targeted adversary attacks as the nominal policy, $\pi_{nom}$, and the RL agent mounting the targeted attack on the nominal policy as the adversarial policy, $\pi_{adv}$. Previous works have demonstrated the vulnerability of RL algorithms to optimization-based action-space perturbations~\cite{lee2020spatiotemporally}. However, such approaches, called "white-box attacks," requires knowledge of $\pi_{nom}$'s architecture and weights for computing the attack. Moreover, while information from $\pi_{nom}$ may be useful for computing untargeted attacks, it is considerably non-trivial to compute a targeted attack via optimization-based approaches. This is because $\pi_{nom}$ only encodes information about optimal actions with respect to the nominal goal. In other words, a query on $\pi_{nom}$ will not reveal information on which action is optimal with respect to some arbitrarily defined adversarial goal. This motivates us to propose an alternative approach where we train a separate RL agent to learn an adversarial policy to predict perturbations which guide $\pi_{nom}$ to an adversarial goal, without requiring knowledge of $\pi_{nom}$'s internal structure, i.e., a black-box attack. To formulate our targeted attack model, we begin with the assumption that the adversary has access to the following:
1) A working copy or surrogate of the trained nominal policy and environment. 2) Nominal policy's input and/or output channels (readings from sensors and actuators). 3) Capability to modify the nominal policies output channel (perturbing the actions)

Next, we consider the objective of the targeted attack model, which is to compute an optimal sequence of perturbations to add to $\pi_{nom}$'s actuation such that $\pi_{nom}$ arrives at an adversarial goal, conditioned only on $\pi_{nom}$'s input/output information. Since we have assumed $\pi_{nom}$ is trained sufficiently and thus can be considered a stationary policy during deployment, the interactions of $\pi_{nom}$ with the environment and corresponding state transitions are essentially another Markov Decision Process (MDP) with stationary dynamics. As such, we can cast the objective of the targeted attack model as another RL formulation, where the interactions of $\pi_{nom}$ with the actual environment are combined into an adversarial training environment for $\pi_{adv}$. In this adversarial environment, the goal of $\pi_{adv}$ is to maximize the reward with respect to the adversarial goal, subject to environment's state transitions and $\pi_{nom}$'s actions. Formally, the objective of the the targeted attack model can be expressed as:
\begin{equation}
\begin{aligned}
& {\text{max}} \  \boldsymbol{R_\text{adv}} = \sum_{t=0}^{T} \hat{R}(\hat{s_t},\delta_{t}) \\
& \text{where :} \\
& \hat{s_t} = \{s_{t}, a_{t}, d_{adv}\}, \ \delta_{t}\sim \pi_{adv}(\hat{s_t})\\
& s_{t} = E(s_{t-1}, a_{t-1} + \delta_{t-1}), \ a_{t}\sim \pi_{nom}(s_t)
\end{aligned}
\end{equation}
In the formulation above, $\hat{s_t}$ denotes the state observed by $\pi_{adv}$ at time step $t$, which consists of concatenated vectors of the nominal state observation $s_{t}$, nominal action $a_{t}$ sampled from $\pi_{nom}$ and a distance measure to the adversary goal $d_{adv,t}$. Additionally, $\delta_{t}$ denotes the perturbations sampled from $\pi_{adv}$ and $E(.)$ signifies state transitions of the environment due to the previous states, actions and perturbations. Without loss of generalizability, the reward function that we used in our experiments for the formulation above is expressed by:

\begin{equation}
\begin{aligned}
& \hat{R} = 
\begin{cases} \|d_{adv, t-1}\| - \|d_{adv, t}\| - I(\lambda) \ \ \text{if} \  d_{adv} > 0\\ 
1 \ \ \ \ \ \ \ \ \ \ \ \ \ \ \ \ \ \ \ \ \ \ \ \ \ \ \ \ \ \ \ \ \ \ \ \ \ \ \ \ \ \ \ \ \ \ \text{if} \ d_{adv} = 0
\end{cases}
\end{aligned}
\label{eqn:reward}
\end{equation}

where I($\lambda$) is a penalty indicator term that penalizes $\pi_{adv}$ if $\pi_{nom}$ reaches the nominal goal. In our experiments, we maintain I($\lambda$) as 1 if the agent is at the nominal goal; otherwise, I($\lambda$) is 0. We highlight that the reward function shown in Equation~\ref{eqn:reward} is just one possible form that works for training  $\pi_{adv}$. Nonetheless, other forms of reward function may be crafted, without affecting the RL formulation of the targeted attack model. With the targeted attack's objective formulated as a RL problem, we can now obtain the attack model, represented by $\pi_{adv}$ via conventional RL training algorithms. A pseudocode for training $\pi_{adv}$ using conventional RL algorithms is also shown in Alg.1 in Supplementary Materials. 

\subsection{Differences from other attack models}

We begin with a brief contrast of our attack model with previously proposed attack models and discuss their respective benefits and drawbacks. In state-space attacks, the attack model perturbs or modifies the input of the RL agent such that the agent is fooled into taking wrong actions. Hence, under the assumption of optimal perturbations, the adversary can consistently trick $\pi_{nom}$ to take actions that eventually lead it to an adversarial goal without any resistance. In contrast, action-space attacks do not aim to fool $\pi_{nom}$ but add perturbations to $\pi_{nom}$'s output such that the resultant actions lead the agent to the adversarial goal. Since the observations of $\pi_{nom}$ are not corrupted and with the assumption that $\pi_{nom} \equiv \pi^{*}$, $\pi_{nom}$ will output an action such that for every $\delta_{t}$ added to $a_{t}$, the action $a_{t+1}$ selected by $\pi_{nom}$ will still be optimal. In other words, a sufficiently trained $\pi_{nom}$ will tend to provide a correcting force at the next time step, which may cancel out the effect of the $\delta_{t}$. This renders the targeted attack model in the action-space more challenging and possibly more limited in efficacy than observation space perturbations.

Another comparison that we would like to discuss is the differences between optimization-based attacks and learning-based attacks. In optimization-based attacks, the optimal perturbations are typically computed via gradient-based optimization strategies. The advantage of these attacks is that they can be computed online, with relatively light compute. However, these methods often require access to internal information of $\pi_{nom}$, such as network architectures and weights, which may be non-trivial to obtain. Additionally, the knowledge encoded within $\pi_{nom}$ typically does not contain accurate information with regard to other arbitrary goals; hence mounting targeted attacks is not straight-forward. On the other hand, our proposed learning-based method circumvents both disadvantages of optimization-based methods. Access to input/output information is significantly less challenging to obtain that $\pi_{nom}$'s network architecture and weights information, hence making such black-box attacks more plausible. Furthermore, by parameterizing the attack model with another goal-conditioned policy, we can learn the optimal perturbations that are directly related to the adversarial goals, which enables us to carry out targeted attacks. Nonetheless, learning-based methods do require additional off-line training time and larger computational resources. The efficacy of learning-based attacks is also highly dependent on the training process, which assumes that the training data distribution, in this case the interactions between $\pi_{nom}$ and the environment, is constant. If $\pi_{nom}$ gets re-trained significantly, we hypothesize that the existing $\pi_{adv}$ would not work very well and will have to be re-trained as well. However, it is still imperative to examine learning-based targeted attacks since the risk of CPS subjected to such attacks exists, and the subsequent effect of being driven to an adversarial state can be costly. 

\section{Results and discussion}

\subsection{Experimental descriptions}
\label{subsec:experimental_description}

To test the efficacy of our attack models, we perform experiments on two platforms, specifically Point-Goal and Car-Goal, found in the suite of environments proposed by~\cite{ray2019benchmarking}. In both environments, the task for the $\pi_{nom}$ is to navigate to a randomly generated goal. To obtain $\pi_{nom}$ as a testing platform for mounting our attacks, we train $\pi_{nom}$ with the PPO algorithm on the two environments for 5E6 steps. After obtaining the trained $\pi_{nom}$, we trained $\pi_{adv}$ to mount the targeted attacks on $\pi_{nom}$. The training of $\pi_{nom}$ was also performed using PPO with 5E6 steps. 

\subsection{Adversarial policy for targeted attack}
\begin{figure}[h]
  \centering
  \includegraphics[width=1\linewidth,  clip, trim={0in 0.5in 0in 0in }]{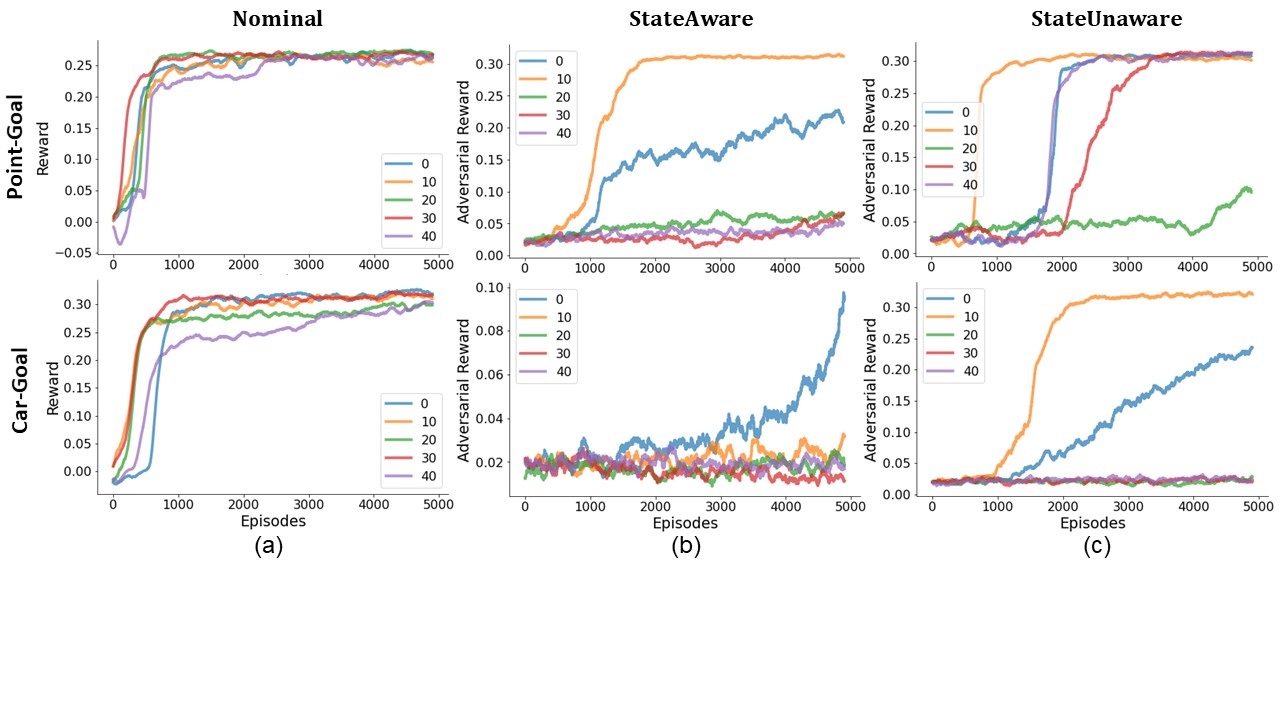}
  \caption{Summary of training rewards for nominal and adversarial policies in Point-Goal and Car-Goal environments. (a) High training rewards for $\pi_{nom}$ illustrates that agents are well-trained in both environments (b) Training rewards for $\pi_{adv}$ that observes $\pi_{nom}$'s state and actions. (c)  Training rewards for $\pi_{adv}$ that only observes $\pi_{nom}$'s actions. Comparing (b) and (c), we observe that perturbations generated by $\pi_{adv}$ that only observes $\pi_{nom}$'s actions are more effective.}
  \label{fig:v4_v5_results}
\end{figure}

In this section, we provide the results of training $\pi_{adv}$ to learn the optimal perturbations, which guides $\pi_{nom}$ towards an adversarial goal for five different seeds in the two different environments. We first demonstrate the rewards of training $\pi_{nom}$ in the first column of Fig.~\ref{fig:v4_v5_results}. As observed, the training of $\pi_{nom}$ converged and is capable of maintaining a high reward across different seeds. Having a sufficiently trained $\pi_{nom}$ allows us to confidently test the effectiveness of $\pi_{adv}$ as an attack model and attribute the $\pi_{nom}$ failures to the efficacy of the proposed attack. 

\begin{table}[h]
\small
  \caption{Mean success rates/episode of $\pi_{adv}$ in perturbing $\pi_{nom}$ to an adversarial goal during evaluation. $\|d_G\|$ denotes the minimum distance an adversarial goal is generated from the nominal goal. As $\|d_G\|$ increases, the adversary success rate generally decreases as it is harder to perturb the $\pi_{nom}$ to the adversarial goal.}
  \label{tab:v4_v5_results}
  \begin{tabular}{l l c c c c}
  \toprule
  && \multicolumn{2}{c}{Point-Goal} &\multicolumn{2}{c}{Car-Goal}\\
  \midrule
  Min. $\|d_{G}\|$ & Goal & $StateAware$ & $StateUnaware$ & $StateAware$ & $StateUnaware$\\
  \midrule
  \multirow{2}{*}{0.5} & Adversary & 6.22 $\pm$ 5.29 & 11.12 $\pm$ 4.52 & 1.46 $\pm$ 1.90 & 5.46 $\pm$ 5.53 \\
                       & Nominal & 2.66 $\pm$ 2.90 & 0.78 $\pm$ 1.08 & 3.62 $\pm$ 3.36 & 4.74 $\pm$ 4.40 \\
  \midrule
  \multirow{2}{*}{1.0} & Adversary & 5.88 $\pm$ 5.26 & 11.28 $\pm$ 4.31 & 1.52 $\pm$ 2.10 & 5.20 $\pm$ 5.76 \\
                       & Nominal & 2.32 $\pm$ 2.85 & 0.50 $\pm$ 1.04 & 3.12 $\pm$ 3.27 & 4.54 $\pm$ 4.22 \\
  \midrule
  \multirow{2}{*}{1.5} & Adversary & 5.30 $\pm$ 5.17 & 11.30 $\pm$ 4.04 & 1.32 $\pm$ 1.75 & 4.76 $\pm$ 4.04 \\
                       & Nominal & 1.84 $\pm$ 2.58 & 0.58 $\pm$ 0.90 & 3.26 $\pm$ 3.42 & 3.92 $\pm$ 5.57 \\
\bottomrule
\end{tabular}
\end{table}
\vspace{-7pt}
Next, we present the training rewards of $\pi_{adv}$ trained in two variants of adversarial environments. The first variant of $\pi_{adv}$, denoted as $StateAware$ in the second column of Fig.~\ref{fig:v4_v5_results}, observes $\pi_{nom}$'s inputs such as position readings, velocity readings, distance to the nominal goal, the actuation outputs, and distance to the adversarial goal. The second variant of $\pi_{adv}$, denoted as $StateUnaware$, observes $\it{only}$ $\pi_{nom}$'s actuation output and distance to the adversarial goal. The training rewards of this variant is shown in the third column of Fig.~\ref{fig:v4_v5_results}. Interestingly, we observed that the $\pi_{adv}$ trained in $StateUnaware$, which consists of a reduced set of observations, achieved higher overall rewards. In the Point-Goal environments, we see that only two of the five $\pi_{adv}$ was able to achieve high rewards in $StateAware$. In contrast, four of the five $\pi_{adv}$ trained in the $StateUnaware$ environment were capable of converging to high rewards. A similar trend is also observed in the Car-Goal environment, which has more complex dynamics. In the $StateAware$ variant of the training environment, only one $\pi_{adv}$ shows signs of learning during the 5 million training steps~\footnote{Codes are available at \url{https://github.com/xylee95/targeted_adversarial_policies}}. However, in the $StateUnaware$ variant, two of the $\pi_{adv}$ showed signs of learning at earlier stages of training and achieved significantly higher rewards.

For validation, we considered three scenarios categorized by the minimum distance an adversarial goal is generated from the nominal goal, denoted by $\|d_{G}\|$. These three scenarios ensure that we evaluate $\pi_{adv}$'s ability on different levels of difficulty, where the adversarial goal may be near or far from the nominal goal. For each scenario, each $\pi_{adv}$ is evaluated for ten episodes, with each episode having a 1,000 steps. A new pair of adversarial and nominal goal at least $\|d_{G}\|$ apart is randomly generated each time $\pi_{nom}$ reaches either the adversarial or nominal goal. The number of time $\pi_{nom}$ reaches either one of the goals is averaged across all episodes and seeds and tabulated in Table~\ref{tab:v4_v5_results}. Overall, we observed that the success rates of $\pi_{adv}$ in generating perturbations that lead $\pi_{nom}$ to the adversarial goal are higher for the Point-Goal environment compared to the Car-Goal environment. This is expected since the dynamics of the Car-Goal environment is more complex and thus harder to compute a perturbation which controls $\pi_{nom}$. However, for both environments and across all scenarios, the $StateUnaware$ variant of $\pi_{adv}$ is significantly more effective in mounting the targeted attacks. Such observation is counter-intuitive since this variant of $\pi_{adv}$ has access to less information on $\pi_{nom}$. This could be attributed to the fact that sufficient information is already embedded in the actions of $\pi_{nom}$ for $\pi_{adv}$ to learn the optimal $\delta_t$. Consequently, additional state information of $\pi_{nom}$ only acts as detrimental noise to $\pi_{adv}$'s learning process. 

Overall, these results suggest that $\pi_{nom}$ trained with conventional RL strategies are susceptible to targeted action-space attacks. More importantly, our results showed that with our proposed attack model, $\pi_{adv}$ only requires access to the $\pi_{nom}$ output (actions) rather than both input and output (state and actions) channels to learn an effective targeted attack. In terms of CPS applications, this is especially crucial since requiring access to only the $\pi_{nom}$ actions lowers the barrier of entry for mounting such targeted attacks. 

\subsection{On robustness of nominal policy}
\begin{figure*}[ht]
    \centering
    \includegraphics[width=\linewidth, clip, trim={0in 4.24in 3.4in 0.6in }]{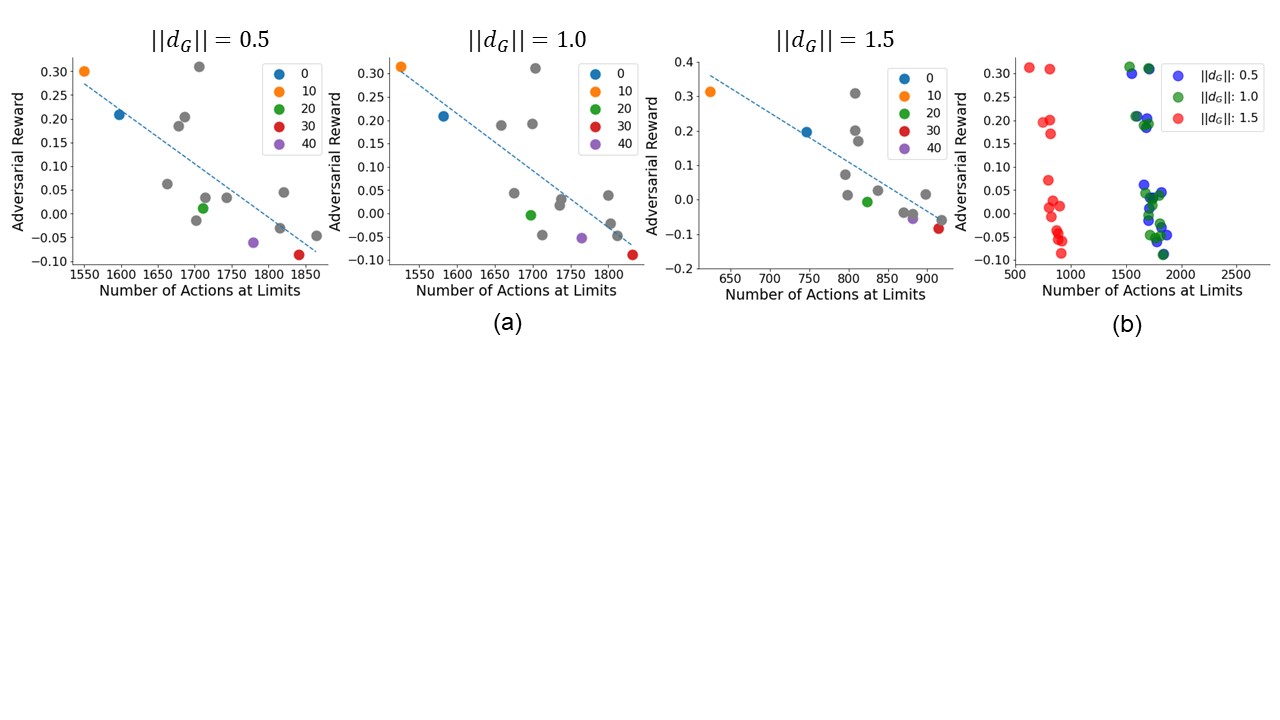}
    \caption{Investigation of different $\pi_{nom}$ behaviors in Car-Goal reveals that a policy that frequently output actions that are at the limits of the entire action space, i.e., 1 or -1, is naturally more robust towards action-space targeted attacks (low adversarial rewards). Colored data points corresponds to original seeds shown in Fig.~\ref{fig:v4_v5_results} and gray data points corresponds to additional random seeds.}
    \label{fig:nominal_robustness}
\end{figure*}
In this section, we provide some insights into what makes a particular policy naturally robust or susceptible to targeted action-space attacks. Having demonstrated that $\pi_{adv}$ that was trained only on the output channels of $\pi_{nom}$ is superior over $\pi_{adv}$ that was trained on the input and output channels, we focus the consequent studies only on the $StateUnaware$ variant of $\pi_{adv}$. As illustrated in the third column of the Car-Goal environment in Fig.~\ref{fig:v4_v5_results}, it is clear that certain trajectories of $\pi_{adv}$ have a more challenging time learning the optimal perturbations to achieve high rewards. To test if this phenomenon is driven by some characteristics of the underlying $\pi_{nom}$, we trained another 15 seeds of $\pi_{nom}$ and $\pi_{adv}$ to observe the training behavior of $\pi_{adv}$. 

In the environments that we used, the range of actions that $\pi_{nom}$ can take is bounded between [-1,1]. We evaluated each $\pi_{nom}$'s behavior under three scenarios of $\|d_G\|$, as shown in Table~\ref{tab:v4_v5_results}, and aggregated the number of actions in both action dimensions that are either -1 or 1 over the entire trajectory of 1,000 steps. In Fig.~\ref{fig:nominal_robustness}, we visualize the final adversarial training rewards of $\pi_{adv}$ as a function of the average number of actions that $\pi_{nom}$ takes that is at the bounds of the action-space. Each data point in the figure represents a unique $\pi_{nom}$ trained on a different seed. As observed, we see a clear negative correlation between the number of actions at the action-space limits with the adversarial reward. A $\pi_{nom}$ that frequently outputs actions that are $\it{inside}$ the range of [-1,1] tend to be more susceptible to targeted attacks, as reflected by higher adversarial rewards. In contrast, a $\pi_{nom}$ that frequently chooses actions that are at the bounds of [-1,1] tend to be more robust to targeted attacks mounted by $\pi_{adv}$. This confirms the hypothesis that certain $\pi_{nom}$ are inherently robust to such targeted adversarial attacks and the common underlying factor among these policies is that they frequently output actions that are either maximum or minimum (similar to a on-off control policy). This is in contrast to previously made observations that identified policies which are more $\it{smooth}$ with respect to state observations are more robust to state space attacks~\cite{russo2019optimal}.

\subsection{Adversarial training schemes}
\begin{figure*}[ht]
    \centering
    \includegraphics[width=\linewidth, clip, trim={0in 4.35in 0in 0.2in}]{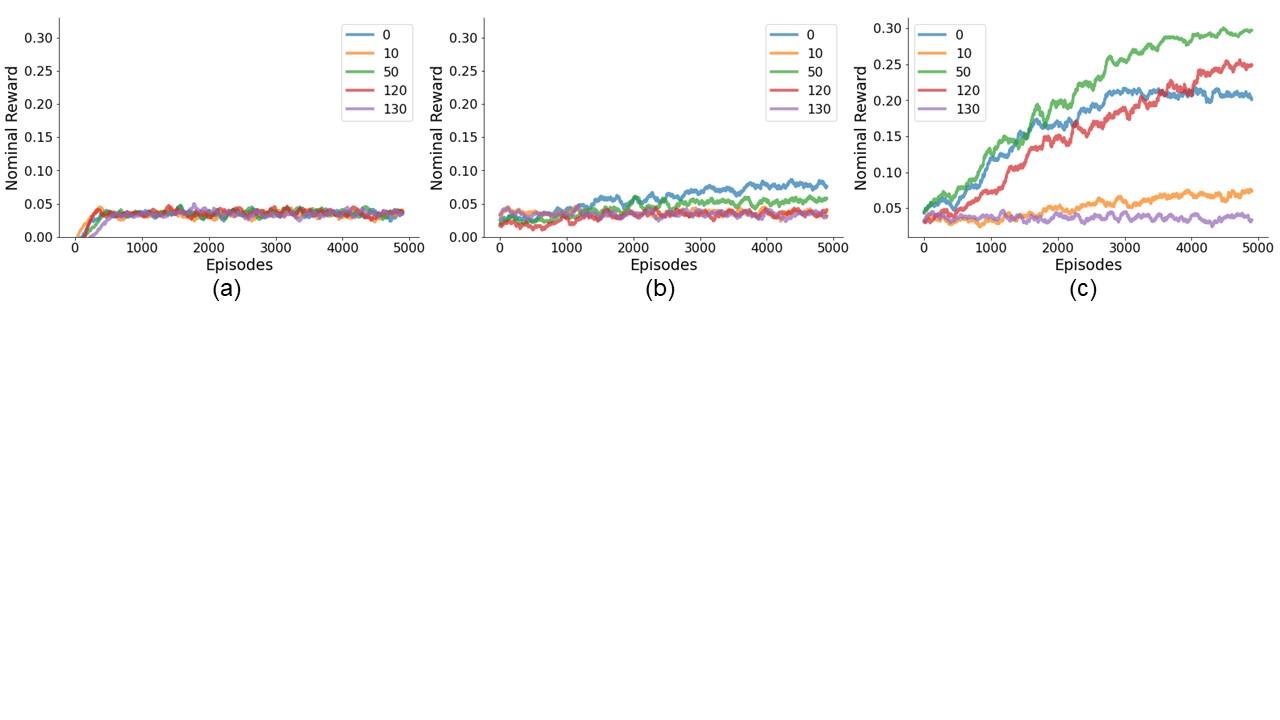}
    \caption{Training curves of $\pi_{nom}$ robustified via adversarial training with (a) untrained $\pi_{nom}$ and $\pi_{adv}$ (b) untrained $\pi_{nom}$ and trained $\pi_{adv}$ and (c) trained $\pi_{nom}$ $\pi_{adv}$. In (c) the weights of the  trained $\pi_{nom}$ were loaded and further fine-tuned by training the agent with targeted attacks. Increasing rewards in (c) reflects the capability of the nominal policy to counter targeted attacks while methods (a) and (b) shows no signs of robustification.}
    \label{fig:TL_training}
\end{figure*}
To induce the behavior of taking extreme actions in $\pi_{nom}$ to be robust to targeted adversarial perturbations, we consider training $\pi_{nom}$ with adversarial training that involves the inclusion of adversarial examples in the training process of the policy. Here, we consider three variants of adversarial training schemes and show that only one variant successfully robustifies $\pi_{nom}$.

In the first variant, we assumed both $\pi_{nom}$ and $\pi_{adv}$ are initially untrained and trained both policies in tandem. In this scheme, $\pi_{nom}$ learns to maximize its reward by reaching the nominal goal, while $\pi_{adv}$ also learns to maximize the adversarial reward by learning the optimal $\delta_t$. This variant of training scheme is ideal since we do not assume access to a trained $\pi_{adv}$ and only require the alternative training of two policies with different objectives. Nonetheless, a drawback of this scheme is that the training process of both policies might not converge since both policies are continually being updated. This also breaks the assumption that the interactions between $\pi_{nom}$ and the environment is an MDP with stationary dynamics in the context of training $\pi_{adv}$. For the second variant, we assume that we have access to a trained $\pi_{adv}$ and train $\pi_{nom}$ with the targeted attacks mounted. In this formulation, $\pi_{adv}$ is fixed, and $\pi_{nom}$ learns a policy that maximizes its reward while being subjected to the perturbations generated by $\pi_{adv}$. Lastly, in the third variant, we assume that both $\pi_{nom}$ and $\pi_{adv}$ are already trained, and we further robustify $\pi_{nom}$ by fine-tuning tits weights via adversarial training. This third variant is similar to the second variant, except that $\pi_{nom}$ isn't trained from scratch but instead trained using a form of transfer learning where the weights of the trained $\pi_{nom}$ are used as a starting point for further fine-tuning. 

In Fig.~\ref{fig:TL_training}, we show the results of robustifying $\pi_{nom}$ via the three variants of adversarial training schemes. As seen in Fig.~\ref{fig:TL_training}(a) and (b), the first two variants failed to robustify $\pi_{nom}$. Nonetheless, the third method of adversarial training, which leverages the prior weights of a pre-trained $\pi_{nom}$ as a starting point, displayed significant learning improvements when compared with the previous methods. We highlight that the rewards of $\pi_{nom}$ shown in Fig.~\ref{fig:TL_training} are the rewards achieved by $\pi_{nom}$ while being subjected to targeted attacks by $\pi_{adv}$. Additionally, only random seeds of $\pi_{noms}$ that were initially vulnerable to targeted attacks were selected for these adversarial training schemes. This ensures that the improvements in $\pi_{nom}$ seen in Fig.~\ref{fig:TL_training}(c) can be solely attributed to the effects of adversarial training. For additional results on the effects of combining transfer learning with adversarial learning to robustify $\pi_{nom}$, please refer to Sec. 2 in the Supplementary Materials.

\section{Conclusions and future work}

In this study, we proposed a learning-based black-box targeted attack model that perturbs the output of a RL agent, $\pi_{nom}$, to lead the agent to an adversarial goal. We show that such an attack model can be represented by another RL agent, $\pi_{adv}$, which learns the optimal perturbations given only the input and/or output channels of $\pi_{nom}$. Experimental results reveal that our proposed attack model is feasible and the $\pi_{adv}$ which only observes the output channel, can mount a  more effective targeted attack. This reveals a vulnerability in systems where the decisions are controlled by a RL agent, even in situations where knowledge $\pi_{nom}$'s weights and architecture are unknown or well-protected. Additionally, we observed that $\pi_{nom}$ are naturally robust when they frequently output action magnitudes that are at the limits of the action range. These results led us to propose a defense scheme that combines adversarial training with transfer learning to induce such behaviors into $\pi_{nom}$ to robustify them against such attacks while maintaining a comparable performance under nominal conditions. The results of training $\pi_{nom}$ via the proposed defense scheme showed that the success rate of $\pi_{adv}$ in driving $\pi_{nom}$ to the adversarial goal decreased by half. Collectively, our experiments illustrate a potential vulnerability in RL agents from targeted attacks in the action-space. Defensive schemes such as adversarial training for $\pi_{nom}$ provide a certain degree of robustness towards such attacks, although it is by no means a fully effective method. 

While we have shown that policies that frequently output extreme actions are more robust towards adversarial attacks and presented one method to induce such behaviors into policies, we acknowledge that such a robustness-inducing mechanism may not be ideal. For one thing, policies that emit such actions closely imitate the behaviors of on-off controllers (bang-bang controllers). In the context of real CPS, such controllers often have negative practical implications such as reduced lifespan of physical parts due to an increase in friction forces, fatigue, and more. Hence, a future direction is to develop methods to robustify the policies towards action-space attacks while maintaining a smoother control behavior. Additionally, while adversarial training does help in robustifying $\pi_{nom}$, the success of robustification is also highly dependent on the random seeds. We hypothesize the failure to robustify certain $\pi_{nom}$ via adversarial training may be because the perturbations generated by $\pi_{adv}$ is too strong that $\pi_{nom}$ fails to obtain a useful learning signal from all the bad trajectories. Thus, future works may investigate advanced adversarial training schemes such as curriculum learning, which increases the severity of attacks on a fixed schedule. 

\section*{Broader Impact}

Our proposed method for mounting targeted action space attacks highlights a broad, existing vulnerability of systems deployed with RL. Although RL is applied in a wide range of applications, we foresee that the impacts of such attacks are especially high in cyber-physical systems in which the decisions of the RL algorithm corresponds to physical actions or actuation. 

While it is clearly unethical for the proposed attacks to be carried out with malicious intents, we hope that highlighting the existence of such attacks to the community will create an awareness. Consequently, this awareness can hopefully spur the development of RL algorithms with built-in safety mechanisms. With that in mind, we believe that researchers involved in ML security and development of new RL algorithms will directly benefit from the results of this study. Additionally, industrial practitioners who are leveraging RL algorithms may also gain a new perspective on the robustness and security of their systems. Nonetheless, it is possible that our proposed methods can be used with malicious intent and we hope that our discussions on robustification of RL algorithms via adversarial training discourage such attempts. Since the data used in the results was purely simulated, we believe the concern of biases in data is not applicable to this study. 

\begin{ack}
This work was supported in part by NSF grant CNS-1845969.
\end{ack}

\bibliographystyle{unsrt}
\bibliography{bib}

\begin{thebibliography}{10}

\bibitem{luo2020deep}
Yuan Luo, Ya~Xiao, Long Cheng, Guojun Peng, and Danfeng~Daphne Yao.
\newblock Deep learning-based anomaly detection in cyber-physical systems:
  Progress and opportunities.
\newblock {\em arXiv preprint arXiv:2003.13213}, 2020.

\bibitem{mozaffari2020learning}
Farnaz~Seyyed Mozaffari, Hadis Karimipour, and Reza~M Parizi.
\newblock Learning based anomaly detection in critical cyber-physical systems.
\newblock In {\em Security of Cyber-Physical Systems}, pages 107--130.
  Springer, 2020.

\bibitem{lee2020automated}
Xian~Yeow Lee, Sourabh~K Saha, Soumik Sarkar, and Brian Giera.
\newblock Automated detection of part quality during two-photon lithography via
  deep learning.
\newblock {\em Additive Manufacturing}, 36:101444, 2020.

\bibitem{lore2016deep}
Kin~Gwn Lore, Nicholas Sweet, Kundan Kumar, Nisar Ahmed, and Soumik Sarkar.
\newblock Deep value of information estimators for collaborative human-machine
  information gathering.
\newblock In {\em 2016 ACM/IEEE 7th International Conference on Cyber-Physical
  Systems (ICCPS)}, pages 1--10. IEEE, 2016.

\bibitem{liu2018deep}
Hongyi Liu, Tongtong Fang, Tianyu Zhou, Yuquan Wang, and Lihui Wang.
\newblock Deep learning-based multimodal control interface for human-robot
  collaboration.
\newblock {\em Procedia CIRP}, 72:3--8, 2018.

\bibitem{popova2018deep}
Mariya Popova, Olexandr Isayev, and Alexander Tropsha.
\newblock Deep reinforcement learning for de novo drug design.
\newblock {\em Science advances}, 4(7):eaap7885, 2018.

\bibitem{lee2019case}
Xian~Yeow Lee, Aditya Balu, Daniel Stoecklein, Baskar Ganapathysubramanian, and
  Soumik Sarkar.
\newblock A case study of deep reinforcement learning for engineering design:
  Application to microfluidic devices for flow sculpting.
\newblock {\em Journal of Mechanical Design}, 141(11), 2019.

\bibitem{jang2019simulation}
Kathy Jang, Eugene Vinitsky, Behdad Chalaki, Ben Remer, Logan Beaver, Andreas~A
  Malikopoulos, and Alexandre Bayen.
\newblock Simulation to scaled city: zero-shot policy transfer for traffic
  control via autonomous vehicles.
\newblock In {\em Proceedings of the 10th ACM/IEEE International Conference on
  Cyber-Physical Systems}, pages 291--300, 2019.

\bibitem{tan2019deep}
Kai~Liang Tan, Subhadipto Poddar, Soumik Sarkar, and Anuj Sharma.
\newblock Deep reinforcement learning for adaptive traffic signal control.
\newblock In {\em Dynamic Systems and Control Conference}, volume 59162, page
  V003T18A006. American Society of Mechanical Engineers, 2019.

\bibitem{10.1145/3178876.3185994}
Guanjie Zheng, Fuzheng Zhang, Zihan Zheng, Yang Xiang, Nicholas~Jing Yuan, Xing
  Xie, and Zhenhui Li.
\newblock Drn: A deep reinforcement learning framework for news recommendation.
\newblock In {\em Proceedings of the 2018 World Wide Web Conference}, WWW '18,
  page 167–176, Republic and Canton of Geneva, CHE, 2018. International World
  Wide Web Conferences Steering Committee.

\bibitem{furuta2019pixelrl}
Ryosuke Furuta, Naoto Inoue, and Toshihiko Yamasaki.
\newblock Pixelrl: Fully convolutional network with reinforcement learning for
  image processing.
\newblock {\em IEEE Transactions on Multimedia}, 2019.

\bibitem{7989385}
S.~{Gu}, E.~{Holly}, T.~{Lillicrap}, and S.~{Levine}.
\newblock Deep reinforcement learning for robotic manipulation with
  asynchronous off-policy updates.
\newblock In {\em 2017 IEEE International Conference on Robotics and Automation
  (ICRA)}, pages 3389--3396, 2017.

\bibitem{7989381}
Y.~{Zhu}, R.~{Mottaghi}, E.~{Kolve}, J.~J. {Lim}, A.~{Gupta}, L.~{Fei-Fei}, and
  A.~{Farhadi}.
\newblock Target-driven visual navigation in indoor scenes using deep
  reinforcement learning.
\newblock In {\em 2017 IEEE International Conference on Robotics and Automation
  (ICRA)}, pages 3357--3364, 2017.

\bibitem{chen2019adversarial}
Tong Chen, Jiqiang Liu, Yingxiao Xiang, Wenjia Niu, Endong Tong, and Zhen Han.
\newblock Adversarial attack and defense in reinforcement learning-from ai
  security view.
\newblock {\em Cybersecurity}, 2(1):11, 2019.

\bibitem{ilahi2020challenges}
Inaam Ilahi, Muhammad Usama, Junaid Qadir, Muhammad~Umar Janjua, Ala Al-Fuqaha,
  Dinh~Thai Hoang, and Dusit Niyato.
\newblock Challenges and countermeasures for adversarial attacks on deep
  reinforcement learning.
\newblock {\em arXiv preprint arXiv:2001.09684}, 2020.

\bibitem{7501620}
S.~{Longhi} and A.~{Monteriù}.
\newblock Fault detection and isolation of linear discrete-time periodic
  systems using the geometric approach.
\newblock {\em IEEE Transactions on Automatic Control}, 62(3):1518--1523, 2017.

\bibitem{basseville1993detection}
Mich{\`e}le Basseville, Igor~V Nikiforov, et~al.
\newblock {\em Detection of abrupt changes: theory and application}, volume
  104.
\newblock prentice Hall Englewood Cliffs, 1993.

\bibitem{vamvoudakis2014detection}
Kyriakos~G Vamvoudakis, Joao~P Hespanha, Bruno Sinopoli, and Yilin Mo.
\newblock Detection in adversarial environments.
\newblock {\em IEEE Transactions on Automatic Control}, 59(12):3209--3223,
  2014.

\bibitem{guo2016optimal}
Ziyang Guo, Dawei Shi, Karl~Henrik Johansson, and Ling Shi.
\newblock Optimal linear cyber-attack on remote state estimation.
\newblock {\em IEEE Transactions on Control of Network Systems}, 4(1):4--13,
  2016.

\bibitem{10.5555/3172077.3172414}
Yen-Chen Lin, Zhang-Wei Hong, Yuan-Hong Liao, Meng-Li Shih, Ming-Yu Liu, and
  Min Sun.
\newblock Tactics of adversarial attack on deep reinforcement learning agents.
\newblock In {\em Proceedings of the 26th International Joint Conference on
  Artificial Intelligence}, IJCAI'17, page 3756–3762. AAAI Press, 2017.

\bibitem{DBLP:journals/corr/BehzadanM17}
Vahid Behzadan and Arslan Munir.
\newblock Vulnerability of deep reinforcement learning to policy induction
  attacks.
\newblock In {\em International Conference on Machine Learning and Data Mining
  in Pattern Recognition}, pages 262--275. Springer, 2017.

\bibitem{s2017adversarial}
Sandy Huang, Nicolas Papernot, Ian Goodfellow, Yan Duan, and Pieter Abbeel.
\newblock Adversarial attacks on neural network policies.
\newblock {\em ICLR Workshop}, 2017.

\bibitem{DBLP:journals/corr/abs-1712-03632}
Anay Pattanaik, Zhenyi Tang, Shuijing Liu, Gautham Bommannan, and Girish
  Chowdhary.
\newblock Robust deep reinforcement learning with adversarial attacks.
\newblock In {\em Proceedings of the 17th International Conference on
  Autonomous Agents and MultiAgent Systems}, pages 2040--2042, 2018.

\bibitem{sun2020stealthy}
Jianwen Sun, Tianwei Zhang, Xiaofei Xie, Lei Ma, Yan Zheng, Kangjie Chen, and
  Yang Liu.
\newblock Stealthy and efficient adversarial attacks against deep reinforcement
  learning.
\newblock In {\em AAAI}, 2020.

\bibitem{mandlekar2017adversarially}
Ajay Mandlekar, Yuke Zhu, Animesh Garg, Li~Fei-Fei, and Silvio Savarese.
\newblock Adversarially robust policy learning: Active construction of
  physically-plausible perturbations.
\newblock In {\em 2017 IEEE/RSJ International Conference on Intelligent Robots
  and Systems (IROS)}, pages 3932--3939. IEEE, 2017.

\bibitem{havens2018online}
Aaron Havens, Zhanhong Jiang, and Soumik Sarkar.
\newblock Online robust policy learning in the presence of unknown adversaries.
\newblock In {\em Advances in Neural Information Processing Systems}, pages
  9916--9926, 2018.

\bibitem{russo2019optimal}
Alessio Russo and Alexandre Proutiere.
\newblock Optimal attacks on reinforcement learning policies.
\newblock {\em arXiv preprint arXiv:1907.13548}, 2019.

\bibitem{zhao2020blackbox}
Yiren Zhao, Ilia Shumailov, Han Cui, Xitong Gao, Robert Mullins, and Ross
  Anderson.
\newblock Blackbox attacks on reinforcement learning agents using approximated
  temporal information.
\newblock In {\em 2020 50th Annual IEEE/IFIP International Conference on
  Dependable Systems and Networks Workshops (DSN-W)}, pages 16--24. IEEE, 2020.

\bibitem{lee2020spatiotemporally}
Xian~Yeow Lee, Sambit Ghadai, Kai~Liang Tan, Chinmay Hegde, and Soumik Sarkar.
\newblock Spatiotemporally constrained action space attacks on deep
  reinforcement learning agents.
\newblock In {\em AAAI}, pages 4577--4584, 2020.

\bibitem{tessler2019action}
Chen Tessler, Yonathan Efroni, and Shie Mannor.
\newblock Action robust reinforcement learning and applications in continuous
  control.
\newblock In {\em International Conference on Machine Learning}, pages
  6215--6224, 2019.

\bibitem{tan2020robustifying}
Kai~Liang Tan, Yasaman Esfandiari, Xian~Yeow Lee, Soumik Sarkar, et~al.
\newblock Robustifying reinforcement learning agents via action space
  adversarial training.
\newblock In {\em 2020 American Control Conference (ACC)}, pages 3959--3964.
  IEEE, 2020.

\bibitem{madry2017towards}
Aleksander Madry, Aleksandar Makelov, Ludwig Schmidt, Dimitris Tsipras, and
  Adrian Vladu.
\newblock Towards deep learning models resistant to adversarial attacks.
\newblock In {\em International Conference on Learning Representations}, 2018.

\bibitem{gleave2019adversarial}
Adam Gleave, Michael Dennis, Cody Wild, Neel Kant, Sergey Levine, and Stuart
  Russell.
\newblock Adversarial policies: Attacking deep reinforcement learning.
\newblock {\em International Conference on Learning Representations}, 2019.

\bibitem{schulman2017proximal}
John Schulman, Filip Wolski, Prafulla Dhariwal, Alec Radford, and Oleg Klimov.
\newblock Proximal policy optimization algorithms.
\newblock {\em arXiv preprint arXiv:1707.06347}, 2017.

\bibitem{mnih2016asynchronous}
Volodymyr Mnih, Adria~Puigdomenech Badia, Mehdi Mirza, Alex Graves, Timothy
  Lillicrap, Tim Harley, David Silver, and Koray Kavukcuoglu.
\newblock Asynchronous methods for deep reinforcement learning.
\newblock In {\em International conference on machine learning}, pages
  1928--1937, 2016.

\bibitem{ray2019benchmarking}
Alex Ray, Joshua Achiam, and Dario Amodei.
\newblock Benchmarking safe exploration in deep reinforcement learning.
\newblock 2019.

\bibitem{fujita2019chainerrl}
Yasuhiro Fujita, Toshiki Kataoka, Prabhat Nagarajan, and Takahiro Ishikawa.
\newblock Chainerrl: A deep reinforcement learning library.
\newblock {\em arXiv preprint arXiv:1912.03905}, 2019.

\end{thebibliography}

\section{Supplementary Materials}
\subsection{Environment Descriptions}

\begin{figure}[h]
  \centering
  \includegraphics[width=0.8\columnwidth, clip, trim={0.2in 1.3in 0.2in 1.2in }]{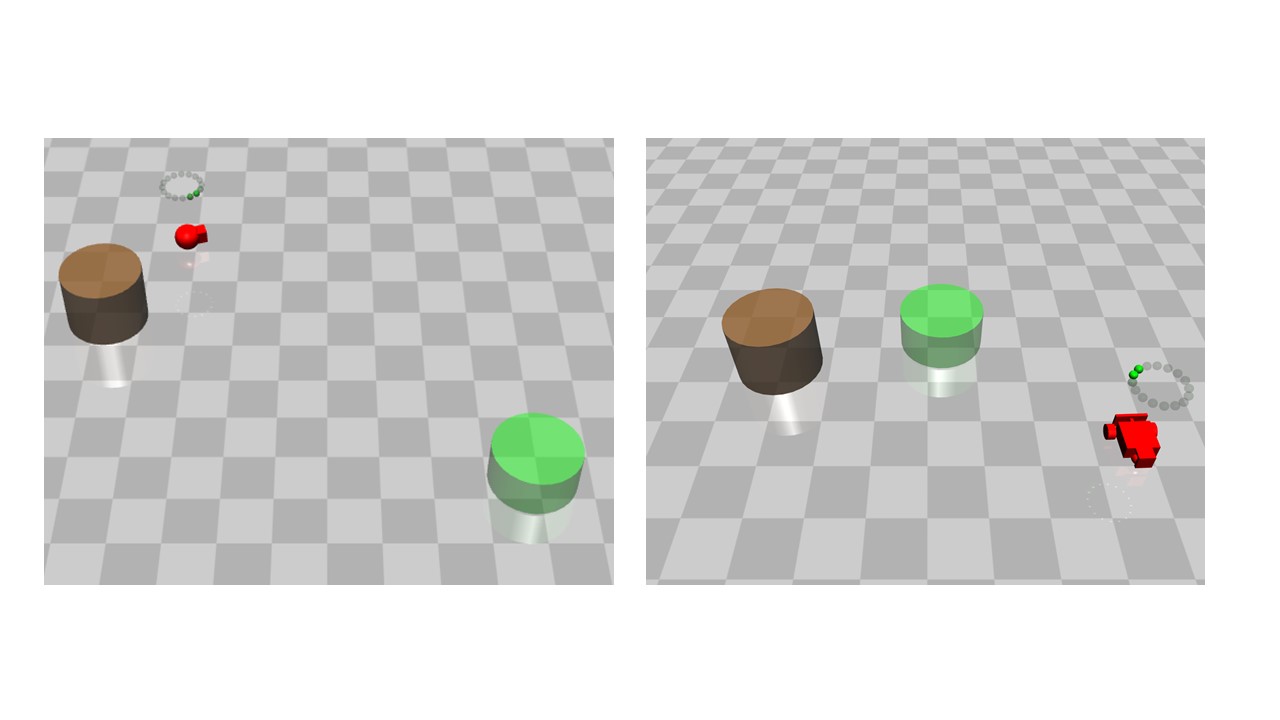}
  \caption{Screenshots of environments used in experiments. Green area, brown area and red object represents the nominal goal, adversarial goal and the agent respectively. Left image shows a birds-eye view of the entire space for Point-Goal environment. Right image shows a zoomed-in view of Car-Goal environment with LiDAR readings visualized as a circular halo above the agent.}
  \label{fig:envs}
\end{figure}

The environments we used as a testing platform is a modified version of the environments proposed by in~\cite{ray2019benchmarking}. The environments are modified such two goals, corresponding to the nominal and adversarial goal are generated for each new trajectory. Fig.~\ref{fig:envs} shows a sample screenshot of each environment. In Point-Goal, the RL agent is represented as a simple robot that can actuate forward/backward and rotate clockwise/anti-clockwise. In Car-Goal, the RL agent is represented by a robot with two-independent wheels and a free-rolling wheel, with the same translation and rotation ability. While both environments have the same action dimensions (translation and rotation), the Car-Goal environment represents a more challenging task with more complex system dynamics. Under nominal conditions, the agent observes its state, which consists of a set of sensor readings and LiDAR information that represents the distance of the agent relative to its goal. The agent then outputs an action bounded between the range of [-1,1] for each actuation dimension.

\subsection{Additional Results}

\begin{table*}[ht]
  \small
  \centering
  \caption{Effects of combining transfer learning with adversarial training on $\pi_{nom}$'s average success rate of reaching nominal and adversarial goals. The success rate of the targeted adversarial attacks on $\pi_{nom}$ decreases by almost half after adversarial training. However, adversarial training also decreases the general performance $\pi_{nom}$ under nominal conditions.}
  \label{tab:TL_results}
  \begin{tabular}{l l c c c c}
  \toprule
  Min.  & Goals & No Adv. Training & With Adv. Training & No Adv. Training &  Adv. Training  \\
  $\|d_{G}\|$ & & with Attack & with Attack & with No Attack  & with No Attack\\

  \midrule
  \multirow{2}{*}{0.5} & Adversary & 10.82$ \pm$ 3.01 & 6.04 $\pm$ 4.89 & N.A. & N.A. \\
                       & Nominal & 0.96 $\pm$ 0.848 & 5.72 $\pm$ 4.61 & 11.5 $\pm$ 1.97 & 9.44 $\pm$ 3.21 \\
  \midrule
  \multirow{2}{*}{1.0} & Adversary & 10.6 $\pm$ 3.35 &  5.68 $\pm$ 4.74 & N.A. & N.A. \\
                       & Nominal & 0.80 $\pm$ 1.10 & 5.48 $\pm$ 4.31 & 11.3 $\pm$ 2.01 & 9.18 $\pm$ 3.12 \\
  \midrule
  \multirow{2}{*}{1.5} & Adversary & 10.1 $\pm$ 3.17 & 5.10 $\pm$ 4.64 & N.A. & N.A. \\
                       & Nominal & 0.66 $\pm$ 0.84 & 5.38 $\pm$ 4.20 & 10.9 $\pm$ 1.57& 8.88 $\pm$ 3.34  \\
  \bottomrule
\end{tabular}
\end{table*}
In this section, we provide additional results of $\pi_{nom}$'s performance pre- and post- adversarial training for different scenarios during evaluation. Comparing $\pi_{nom}$ without adversarial training and $\pi_{nom}$ after adversarial training that is subjected to targeted attacks, we observe that the number of adversarial goals reached decreases almost by half, while the number of nominal goals reached also increased significantly. This illustrates the effectiveness of an adversarial training scheme in providing a certain degree of robustness to $\pi_{nom}$. Nevertheless, such adversarial training schemes also comes with a cost to the general performance of $\pi_{nom}$. Comparing the nominal performance (i.e., performance without attacks) of $\pi_{nom}$ with and without adversarial training, we note that the number of nominal goals reached is slightly lower for $\pi_{nom}$ that has been adversarially trained. However, similar observations have also been made with adversarial training schemes for RL in the state/observation space~\cite{tan2020robustifying} as well as with supervised deep learning tasks~\cite{madry2017towards}. This reveals a trade-off between a policy's performance and ability to generalize to more scenarios, such as adversarial situations. 

\begin{figure}[h]
    \centering
    \includegraphics[width=0.8\linewidth, clip, trim={0in 2.65in 1.1in 0.0in }]{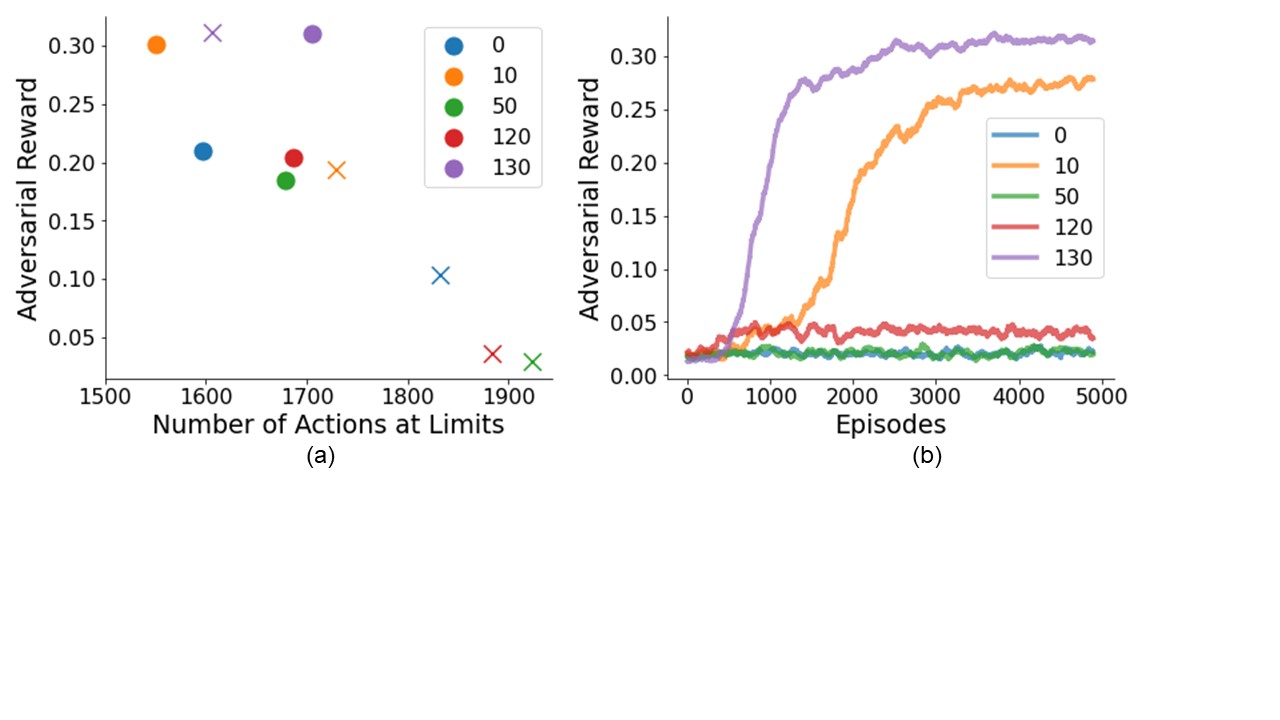}
    \caption{(a) Effects of adversarial training on $\pi_{nom}$'s behaviour. Circles denote $\pi_{nom}$ before adversarial training and crosses denote the same set of $\pi_{nom}$ after adversarial training. The average number of extreme actions that $\pi_{nom}$ takes in a single trajectory increased significantly after adversarial training for seeds that were successfully robustified. (b) Training curves of new $\pi_{adv}$ for attacking robustified $\pi_{nom}$ that were previously vulnerable to targeted attacks. Observe that once $\pi_{nom}$ are successfully robustified, as shown in Fig.4(c) of the main manuscript for seeds (0, 50, 120), it is difficult to learn new $\pi_{adv}$ that effectively mounts the attacks. }
    \label{fig:attackability}
\end{figure}

Next, we show some anecdotal observations on the mechanisms of robustifying $\pi_{nom}$ via adversarial training. In Fig.~\ref{fig:attackability}(a), we visualize the average number of extreme actions that $\pi_{nom}$ outputs pre-adversarial training (in circular markers) and post-adversarial training (in cross markers) for the five seeds shown in Fig.4(c) of the main manuscript. We observe that the number of extreme actions chosen by $\pi_{nom}$ increased significantly for the three seeds that achieved high rewards during adversarial training. These results highlight the role of adversarial training as one possible method for obtaining such policies. Furthermore, this also validates our previous conclusion that a policy that frequently outputs extreme actions is naturally more robust towards targeted attacks. For the sake of completeness, we also tried training new $\pi_{advs}$ to mount targeted attacks on the robustified $\pi_{nom}$ that has been adversarially trained. The results, shown in Fig.~\ref{fig:attackability}(b), reveals that $\pi_{nom}$ that shows successful signs of robustification during adversarial training (seeds 0, 50, 120) remains robust in the presence of new $\pi_{adv}$. Nonetheless, $\pi_{nom}$ that did not get robustified during adversarial training remains vulnerable to new $\pi_{adv}$. 

\subsection{Training algorithm and details}

\begin{algorithm}[h]
\SetAlgoLined
 Initialize $\pi_{adv}$, trained $\pi_{nom}$, $E$\, $N$;
 \While{steps < $N$}{
  Initialize $s_t$\;
  \While{$t$ < $T$}{
   $a_{t}$ $\sim$ $\pi_{nom}(s_t)$\;
   Construct $\hat{s_t}$ = $(s_{t},a_{t}, d_{adv})$ or $\hat{s_t}$ = $(a_{t}, d_{adv})$\;
   $\delta_t$ $\sim$ $\pi_{adv}$ $(\hat{s_t})$\;
   $s_{t+1}$ = $E(s_{t}, a_{t} + \delta_{t})$
   }
  \If{time for update}{
    Update $\pi_{adv}$
    }
 }
 \caption{Algorithm for training $\pi_{adv}$. }
 \label{alg:pi_adv}
\end{algorithm}

\begin{table}[h]
  \centering
  \caption{Parameters used to train $\pi_{nom}$ and $\pi_{adv}$.}
  \label{tab:params}
  \begin{tabular}{l c}
    \toprule
     Parameter & Value  \\ 
     \midrule
     Optimizer      & Adam  \\  
     Learning Rate  & 3E-4  \\
     Entropy Coefficient & 0 \\
     Batch Size & 1024 \\
     Update Interval & 2048 \\
     \bottomrule
\end{tabular}
\end{table}

Relevant details for training the $\pi_{nom}$ and $\pi_{adv}$ are provided in this section. To train the $\pi_{nom}$, we parameterize the agent with two deep neural networks, one representing the policy and one representing the value function. The policy networks consists of three fully-connected layers with the first and second layer having 64 hidden units. The output of the policy network is parameterized with a Gaussian distribution where the mean and variance of the distribution being estimated by the final fully-connected layer with the same dimensions as the action-space. The value function network also consists of three fully connected layers with 128, 64 and 1 hidden units respectively, with the output of the final layer being an estimate of the state's value. A hyperbolic tangent activation function is used in between all layers. The same network architecture was also used to train $\pi_{adv}$ as well. Both $\pi_{nom}$ and $\pi_{adv}$ were trained using PPO with the parameters shown in Table~\ref{tab:params}. All experiments and codes were implemented in ChainerRL~\cite{fujita2019chainerrl}.

\end{document}